\documentclass[letterpaper, 10 pt, conference]{ieeeconf}  % Comment this line out if you need a4paper

\usepackage{pdfpages}
\usepackage{amssymb}
\usepackage{graphicx}
\usepackage{amsmath}
\usepackage{mathrsfs}
\usepackage{subfigure}
\usepackage{wrapfig}
\usepackage{caption}
\usepackage{hyperref}
\usepackage[normalem]{ulem}
\usepackage{ifthen}
\usepackage{color}
\definecolor{gray}{rgb}{0.5,0.5,0.5}
\definecolor{green}{rgb}{0, 0.4, 0}
\definecolor{orange}{rgb}{1, 0.5, 0}
\definecolor{mahogany}{rgb}{0.75, 0.25, 0.0}
\definecolor{purple}{rgb}{0.6, 0, 0.6}
\definecolor{darkgreen}{rgb}{0, 0.4, 0}
\definecolor{frenchblue}{rgb}{0.0, 0.45, 0.73}
\definecolor{red}{rgb}{1,0,0}
\definecolor{yellow}{rgb}{1,1,0}
\newboolean{revising}
\setboolean{revising}{false}
\ifthenelse{\boolean{revising}}
{
	\newcommand{\ignore}[1]{}

    \newcommand{\juanting}[1]{{\color{blue}{#1}}}
	
    \newcommand{\johnson}[1]{{\color{green}{#1}}}

    \newcommand{\todo}[1]{{\color{red}{#1}}}
	
    \newcommand{\wtf}[1]{{\color{yellow}{#1}}}
	
} {
	\newcommand{\ignore}[1]{}

    \newcommand{\juanting}[1]{#1}
	
    \newcommand{\johnson}[1]{#1}

    \newcommand{\todo}[1]{#1}
	
    \newcommand{\wtf}[1]{#1}
	
}

\IEEEoverridecommandlockouts                              % This command is only needed if 
\title{\LARGE \bf
Omnidirectional CNN for Visual Place Recognition and Navigation
}

\author{Tsun-Hsuan Wang$^{*1}$, Hung-Jui Huang$^{*2}$, Juan-Ting Lin$^{1}$, Chan-Wei Hu$^{1}$, Kuo-Hao Zeng$^{1}$, Min Sun$^{1}$
\thanks{$^{*}$ The first two authors contributed equally to the paper. {\tt\small \{johnsonwang0810, stud30821\}@gmail.com}}
\thanks{$^{1}$ National Tsing Hua University}
\thanks{$^{2}$ Massachusetts Institute of Technology}
}

\begin{document}

\maketitle
\thispagestyle{empty}
\pagestyle{empty}

%%%%%%%%%%%%%%%%%%%%%%%%%%%%%%%%%%%%%%%%%%%%%%%%%%%%%%%%%%%%%%%%%%%%%%%%%%%%%%%%
\begin{abstract}
Visual place recognition is challenging, especially when only a few place exemplars are given. To mitigate the challenge, we consider place recognition method using omnidirectional cameras and propose a novel Omnidirectional Convolutional Neural Network (O-CNN) to handle severe camera pose variation.
Given a visual input, the task of the O-CNN is not to retrieve the matched place exemplar, but to retrieve the closest place exemplar and estimate the relative distance between the input and the closest place.
With the ability to estimate relative distance, a heuristic policy is proposed to navigate a robot to the retrieved closest place.
Note that the network is designed to take advantage of the omnidirectional view by incorporating circular padding and rotation invariance. To train a powerful O-CNN, we build a virtual world for training on a large scale. We also propose a continuous lifted structured feature embedding loss to learn the concept of distance efficiently. Finally, our experimental results confirm that our method achieves state-of-the-art accuracy and speed with both the virtual world and real-world datasets.

\end{abstract}

\section{Introduction}

Visual place recognition is a critical task for robot navigation.
For instance, a delivering robot needs to recognize its place (used interchangeably with location) in order to deliver goods to a specific destination.
When many place exemplars (i.e., map consisting of images of places) are given, the main challenge in place recognition is how to efficiently retrieve matched exemplars~\cite{disloc2014,retrieval_cite2_vocab_tree,retrieval_cite1}. On the other hand, when only a few place exemplars are given, many robot visual inputs are away from the exemplars on the map.
In order to localize the robot onto the map, the main challenge becomes how to find and navigate to the ``closest place exemplar" given significantly different camera poses between visual inputs and exemplars. 
Although a few learning-based methods have been proposed to improve the robustness of retrieving exemplars under large camera pose variation \cite{NetVLAD}, they typically do not address how to navigate to the closest place exemplar.

%There are two main schemes for visual navigation: map-based and map-less navigations. Map-based methods {\cite{map-based0,map-based,map-based2,map-based3}} typically rely on robust mapping and localization techniques. However, map-based methods can fail catastrophically when one of the technique failed. For map-less navigation, a few deep learning based methods {\cite{zhu2016target,CMP}} have been proposed to possess implicit knowledge of the environment for reaching their goals. Nevertheless, most methods do not generalize well in unseen environments. 
%As a result, neither scheme alone is robust enough to solve the visual navigation task completely.

\begin{figure}[t!]
	\centering
		\includegraphics[width=0.5\textwidth]{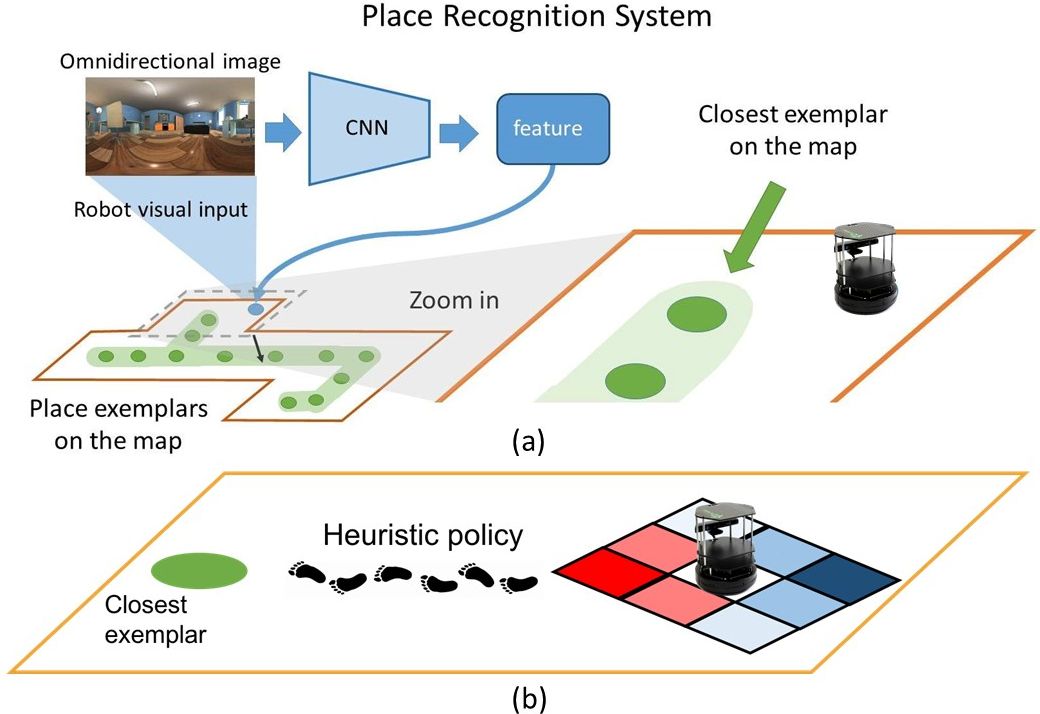}
	\caption{\small \textbf{Overview}. (a) Our visual place recognition system takes an omnidirectional visual input from a robot and, then, retrieve the closest place exemplar on the map using our O-CNN. (b) O-CNN is further used to help navigate a robot to the closest place.}
    \vspace{-7mm}
	\label{fig:abstract}
\end{figure}

%We believe that the key to a robust visual navigation system lies in bridging the gap between these two schemes. 
%In particular, we consider the following scenario. A map is given to a robot, but the robot failed to localize itself in the map and is currently in a map-less condition. Instead of fixing the map-less condition and completely ignore the map, 
In this paper, we consider a robot that is only given a few place exemplars  (i.e., a map with few places), and its current location is unknown and away from these exemplars. Our goal is to retrieve the closest place exemplar and navigate to the \todo{\johnson{closest place, }}for which there is an exemplar (see Fig. ~\ref{fig:abstract}).
We propose to learn a function to estimate the distance between a visual input and a place exemplar captured at \todo{\johnson{two different places}}. Using this function, we can estimate the closest place on the map to the robots current location. We further propose a heuristic policy to take actions to \todo{\johnson{reduce the estimated distance}} from the new robot location to the closest place on the map.

Our main technical contribution lies in how to learn a robust distance function. To do so, we take advantage of both a modern omnidirectional camera configuration and deep learning techniques with our CNN model. On the one hand, omnidirectional cameras have become widely available on the consumer market. The literature has shown that omnidirectional camera can improve the quality of both mapping \cite{kim2003slam,scaramuzza2008appearance} and localization \cite{topological_cite1}. On the other hand, \todo{\johnson{CNN-based (Convolutional Neural Networks) methods have been proposed }}% Chanwei
to estimate the camera pose~\cite{posenet}, recognize place information~\cite{NetVLAD}, etc. Combining these two ideas, we propose a novel omnidirectional Convolutional Neural Network (O-CNN) which is trained completely in virtual environments. The virtual world offers us label-free and agent-safe experimental environments.

Our O-CNN has three important design elements. Firstly, we apply circular padding to both image and CNN feature spaces to reflect the fact that \todo{\johnson{omnidirectional images have}} no true image boundary. Secondly, we propose a novel roll branching approach to conquer rotational variation in the captured omnidirectional images. 
Finally, we modify the lifted structured feature embedding \cite{songCVPR16} to offer the model the concept of distance in the environments, which we call it continuous lifted structured feature embedding.

We show in experimental results that our method is more effective and more efficient in indoor place recognition than several strong baselines in both virtual world and real-world data. Finally, combining our method with our proposed heuristic navigation policy, our whole system provides a joint place recognition and navigation solution.

To sum up, we made the following contributions: (1) we are the first one using a novel CNN model designed for omnidirectional visual input (O-CNN) to address indoor place recognition problems, (2) our O-CNN consists of three unique designs to improve the performance, and (3) we constructed an indoor virtual world and real world dataset for further research use. The final experimental results confirm that our proposed method achieves state-of-the-art accuracy and speed with both the virtual world and real-world datasets.

\section{Related Work}
%{\sout{In this work, we aim to bridge the gap from map-less to map-based navigation in an unseen indoor environment by combining with two active research fields: (1) localization and navigation, (2) deep metric learning. We discuss such two research fields as follows.}}

%{\color{blue}In this work, we aim to propose a efficacious and efficient way to improve the pose invariance in place recognition and further demonstrate that the method can be used in navigation system by adopting the omni-directional perception and topological metric map construction. In the following section, we discuss two active research fields: (1) place recognition and localization; (2) deep metric learning. }

%In this work, while doing place recognition in a sparse topological metric map, we propose an efficacious and efficient way to deal with pose invariance issue and improve the recall rate by adopting omnidirectional perception and deep metric learning. We further demonstrate our place recognition method on naive navigation system.  
%In the following section, 
We discuss three research fields: (1) descriptors for visual place recognition; (2) visual place recognition and navigation; (3) deep metric learning.

\subsection{Descriptors for Visual Place Recognition}
There are two types of visual descriptors for place recognition: (1) Local feature descriptors (2) Global feature descriptors.
%The image processing modules in visual place recognition system can be divided into two main category: (1) Local feature descriptor (2) Global feature descriptor. 
Local feature descriptors describe the image only at noticeable parts. Such methods include SIFT~\cite{local_descriptor_cite1}, SURF~\cite{local_descriptor_cite2}, ORB~\cite{local_descriptor_cite3}, which was recently used in ORB-SLAM system~\cite{orb_slam}, etc.
On the other hand, global descriptors describe the whole image by using histogram and principle component analysis on corners~\cite{global_descriptor_cite2}, edges~\cite{global_descriptor_cite1_edge}, colors~\cite{topological_cite1}, etc. Methods extracting local feature descriptors uniformly from the whole image can be considered one kind of global descriptor as well~\cite{global_descriptor_cite3}.
Convolution Neural Networks have recently been used as learned global feature extractors. Several pieces of research have demonstrated the ability of CNNs to handle appearance variation in changing environment and camera pose variation~\cite{cnn_descriptor_cite1, cnn_descriptor_cite2}.

%\cite{local_descriptor_cite4, local_descriptor_cite5}

\subsection{Visual Place Recognition and Navigation} 

Visual place recognition and navigation are two highly related fields~\cite{visual_place_recogntion_survey}. Place recognition systems usually contain maps that serve to record visited places with different level of abstraction and/or metric information. 
The most simple mapping framework is pure image retrieval \cite{retrieval_cite1}. %These methods is highly similar to the image retrieval tasks in computer vision which distinguish images simply from appearance differences.
Some methods such as hierarchical vocabulary trees \cite{retrieval_cite2_vocab_tree}, adopted by Schindler {\it et al} in \cite{retrieval_cite1}, can improve efficiency when the scale of the map is large.
Some place recognition methods directly regress the 6-DoF camera pose from the observation to obtain absolute position~\cite{brachmann2016dsac,hazirbasimage,kendall2016modelling,kendall2017posenet,posenet}. These methods typically assume that place exemplars on the map are abundant and nicely cover the space of visual inputs.

To increase the efficiency of the matching process, some methods use topological maps to record the relative positions between visited locations \cite{topological_cite2_fabmap, topological_cite3_seqslam, topological_cite4, topological_cite1}. With this method, when the system matches the current location with the visited locations, it only needs to consider the places near the current location \cite{topological_cite8, topological_cite6_cat_slam, topological_cite7,topological_cite5}.
The topological map based methods can be further enhanced by incorporating metric information such as distance and orientation between different nodes on the topological map \cite{topological_cite6_cat_slam, topological_metric_cite1_smart}. The included metric information ranges from sparse landmark maps \cite{ topological_metric_cite3_sparse, topological_metric_cite4_sparse,topological_metric_cite2_sparse} to dense grid maps  \cite{topological_metric_cite5_dense}.
These methods are orthogonal to our proposed method since we assume the visual input is distant from place exemplars on the map.

Given a topological map, robot navigation can be abstractly done by following edges connecting nodes in the map. Furthermore, robot navigation can be more precise with dense metric data.
%Moreover, with the additional topological information, the recognition process can be performed on images with lower resolution to mitigate the memory consumption \cite{topological_cite9_lowRes, topological_cite10_lowRes}.
Map-less navigation methods are also common {\cite{map-less1,map-less2}. These methods tackled obstacle avoidance with visual observation. Other relevant works focus on using only visual observation for navigation tasks {\cite{CMP,zhu2016target}.}}
In our scenario, we are in between mapped and map-less navigation. Given a retrieved closest place, we need to navigate to the closest place without the use of map information. Once, the closest place is reached, we can apply other map-based navigation approaches.

\subsection{Deep Metric Learning}
Deep metric learning plays an important role in this work of mapping appearance difference to spatial metric distance between the robot's visual input and exemplars on a map.
Some previous works present algorithms that learn the representations to estimate the difference between data samples {\cite{NIPS1993_769, Chechik,Chopra}}.
Hu et al. {\cite{Hu}} present a method to learn a set of hierarchical nonlinear transformations to project face pairs into the same feature subspace.
Hoffer et al. {\cite{DBLP}} train a triplet network to learn the representations using distance comparisons.
Song et al. {\cite{songCVPR16}} propose an algorithm for training batches in the neural network via lifted structured feature embedding.
Inspired by these previous works, we propose continuous lifted structure feature embedding which learns representations that can robustly estimate real-world relative distances from a pair of omnidirectional visual inputs.

\section{Method}
\label{sec:method}

Our main contribution is the O-CNN, which extracts deep features form omnidirectional images for visual place recognition. In this section, we first define our visual place recognition system. Then, O-CNN is further described. Finally, a heuristic navigation policy is introduced to show the effectiveness of our O-CNN\johnson{-}based system.

\subsection{Overview}
Given a map consisting of a few place exemplar images, our visual place recognition system is designed to retrieve the closest place exemplar on the map with respect to a robot and to help the robot navigate to the closest place. 
%Topological map is built following the method proposed in \cite{Ko2016ASD}. Initially, the robot walks around the environment and takes omnidirectional photos in a certain frequency. 
We encode both the exemplar images and robot visual input using a whole-image descriptor/feature extracted from the O-CNN.
%is stored for later-on place recognition.
The descriptor of the robot input is compared to all stored descriptors of exemplars on the map by computing feature distance. The minimal feature distance indicates the closest place exemplar \johnson{in terms of real-world location}. In addition, as the robot is getting closer to the place exemplar, the distance between the descriptor of current visual input and the descriptor of the place exemplar will decrease. This information is used to aid the robot's navigation.
    
\begin{figure*}[t]
\centering
\vspace{1mm}
\includegraphics[width=0.8\textwidth]{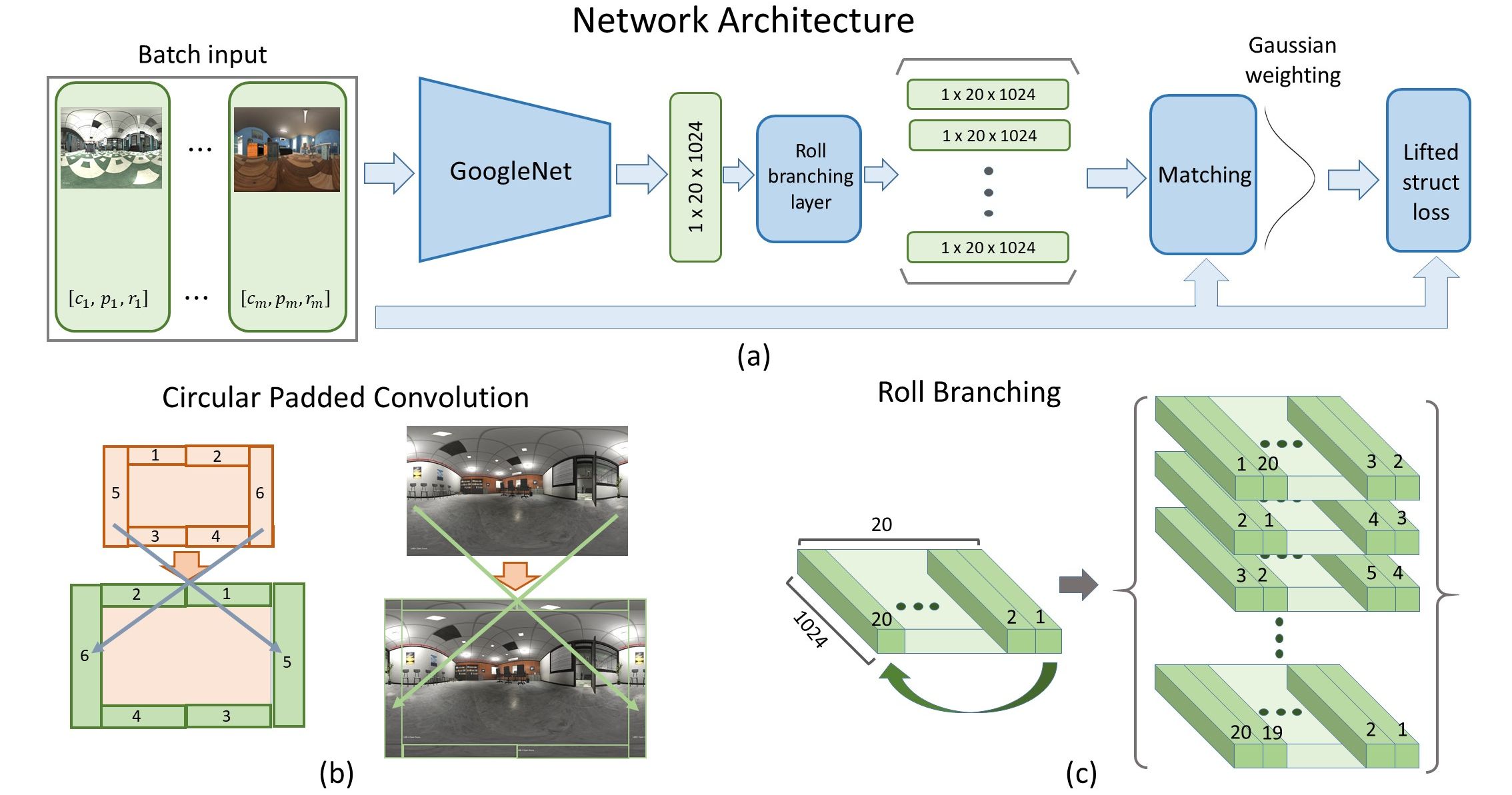}
\caption{\small \textbf{Network Architecture}. (a) In our O-CNN, we add circular padding to each convolution operation in the GoogleNet. After feature extraction, we further perform roll branching to make the architecture robust to purely perspective rotation and compute the lifted structure embedding loss for training. (b) The illustration of the circular convolution operation. We take omnidirectional image as an example, but we actually perform the operation on feature maps after every convolution layer. (c) The illustration of roll branching: After roll branching, we have 20x shifted feature map.}
\vspace{-5mm}
\label{fig:lift_loss}
\end{figure*}

\subsection{O-CNN Architecture}
Input of O-CNN is an omnidirectional image and output of O-CNN is a $w \times d$ feature map, where $w$ is the number of discretized rotations, e.g. if $w=20$, each discretized rotation will cover $360/20=18$ degree of rotation, and $d$ is the dimension of a feature vector of a part of the image specified by the rotation. We introduce three key components to improve performance in our method. Two of them deal with omnidirectional images, and the last one is a loss function adapted from lifted structure metric learning {\cite{songCVPR16}}. In the training stage, each omnidirectional image $I_{i}$ is paired with a room label {\it $c_i$}, a position label {\it $p_i$}, and a rotation label {\it $r_i$}. Data are paired based on {\it $c_i$} and {\it $p_i$}, which will be further explained in Equation (2). An illustration of the whole architecture is in Fig. \ref{fig:lift_loss}(a).
	
	\subsubsection{Circular Padding - Efficiently Using All Information In Omnidirectional image}
    Circular padding is a padding method specifically designed for omnidirectional images in order to eliminate loss of information while adopting other padding methods such as zero or same padding. 
For circular padding, in the horizontal axis, we pad the left of the image by the rightmost part of the image, and vice versa; in the vertical axis, We pad the upper left of the image by the upper right, and vice versa.
    %we pad the left of the image by the right most part of the image and the right of the image by left most part of the image.
    %We further pad upper left of the image by upper right and upper right by upper left, same to the bottom padding.
\todo{\johnson{We apply circular padding for every convolution layer since padding in feature space brings slightly better performance in experiments.}} % Juan
An illustration of circular padding can be found in Fig. \ref{fig:lift_loss}(b). %Circular padding may be used in various other equirectangular omnidirectional image based deep neural network. 

	\subsubsection{Roll Branching - Omnidirectional Camera Rotational Invariance}
    To deal with conditions where two omnidirectional images are taken in the same location with different viewing angles and their relative positions in feature space should be close to each other, we take all discretized horizontal rotations into consideration. Given an input feature map $z^{l}$ of shape ${h}\times{w}\times{d}$, roll branching layer outputs $\mathbf{y^{l}}=\{z_{0}^{l},z_{1}^{l},...,z_{w-1}^{l}\}$, where $z_{k}^{l}$ is $z^{l}$ shifted left by $k$ in the ${w}$ dimension. During inference, given extracted features from image $\mathbf{y^{l}}(I_{i})$ and $\mathbf{y^{l}}(I_{j})$, we define rolling metric distance $\hat{R_{ij}}=\{||z_{k}^{l}(I_{i})-z_{0}^{l}(I_{j})||_2 \mid k\in[0,w-1]\}$. Feature distance $D_{ij}$ between $I_{i}$ and $I_{j}$ is $\min_{k}\hat{R_{ij}}$ and relative rotation estimate $\hat{r_{ij}}=\arg\min_{k}\hat{R_{ij}}$. \johnson{During training, relative rotation ground truth $r_{ij}$ can be computed using rotation labels $r_{i}$ and $r_{j}$, and the loss for positive examples is computed using $\hat{R_{ij}}$ masked by a gaussian distribution centered at $r_{ij}$ (see $g(r_{ij},w)$ in Equation (2)), i.e. two images in a positive pair are rotated to the same orientation before computing loss.} An illustration of roll branching can be found in Fig. \ref{fig:lift_loss}(c).
    \subsubsection{Continuous Lifted Structure Feature Embedding - Mapping Feature Difference to 3D Space Difference}
	\johnson{Our task has a subtle but critical difference from conventional metric learning. Conventional metric learning aims to map inputs to feature space where similarity of inputs can be estimated using distance, but this similarity is defined over discrete labels, e.g. kitchen, toilet. Our task poses a more challenging scenario in which we need to rank difference between continuous position labels. For example, given a fixed target, features with smaller distance in terms of real-world position from the target should be closer to the target feature in the feature space. Our continuous lifted structured loss is adapted from a powerful deep metric learning, lifted loss \cite{songCVPR16} $J_{ij,ori}$, see Equation (3). The original lifted loss mines all negative pairs with respect to both examples per each positive pairs. It then contracts the positive pair and at the same time pulls apart all negative pairs. In traditional metric learning, the definition of positive pairs is fixed, e.g. two images both taken in the bathroom always form a positive pair, whereas, in continuous lifted loss, the positiveness of a pair is not static, i.e. two images can sometimes be a positive pair and sometimes be a negative pair. Continuous lifted loss $J$ is shown in Equation (1) and (2).
    
    \small
    \vspace{1mm}
    \begin{equation}
    J = \frac{1}{2|\hat{P}|}\sum_{(i,j)\in{\hat{P}}}\max(0, J_{ij})^2
    \end{equation}
    \begin{equation}
    \begin{aligned}
    J_{ij} = \log(\sum_{(i,k)\in{(\hat{N}+\hat{PN_i})}}\sum_{D_{ik}^{w}\in{R_{ik}}}\exp(\alpha - D_{ik}^{w}) +\\ \sum_{(j,l)\in{(\hat{N}+\hat{PN_j})}}\sum_{D_{jl}^{w}\in{R_{jl}}}\exp(\alpha - D_{jl}^{w})) + \sum_{D_{ij}^{w}\in{R_{ij}}}g(r_{ij},w)D_{ij}^{w}
    \end{aligned}
    \end{equation}
    \normalsize
    \footnotesize
    \begin{equation}
    \begin{aligned}
    J_{ij,ori} = \log(\sum_{(i,k)\in{N}}\exp(\alpha - D_{ik}) + \sum_{(j,l)\in{N}}\exp(\alpha - D_{jl})) + D_{i,j}~,
    \end{aligned}
    \end{equation}
    \normalsize
    where $\hat{P}$ is all pair set, $\hat{N}$ is negative pair set, $\hat{PN}$ is \todo{\johnson{pseudo-negative}} pair set, $D_{ij}^{w}$ is the distance between feature $i$ and $j$ in the $w^{th}$ branch, and $g(r_{ij},w)$ is the $w^{th}$ bin of a gaussian distribution centered at $r_{ij}$. Each pair in $\hat{N}$ has different room labels and each pair in $\hat{PN}$ has the same room label. We define $d_{ij}$ to be the position distance between two exemplars $i$ and $j$ in physical world. We first randomly choose a pair of points with the same room label $c_{i}$ to be a positive pair $(i,j)$, and a \todo{\johnson{pseudo-negative}} pair consists of a point in the positive pair ($i$ here for illustration) and one of all other points in the same room, where position distance in the \todo{\johnson{pseudo-negative}} pair should be larger than that of the positive pair $d_{ij}$, i.e. $\hat{PN_i}(j) = \{(i,k)|k\in{\hat{P}}, d_{ik}\geq{d_{ij}}\}$. Each time a positive pair is sampled and contracted, all negative and pseudo-negative pairs are pulled apart once. Because positive pairs are defined using position distance, which results in its non-staticness, and the pseudo-negative pairs are obtained depending on positive pairs, pairs with larger position distances will be picked more times to construct $\hat{PN}$. Accordingly, pairs with larger position distances will be pulled apart in feature space more times than those with smaller position distances. Finally, correspondence between feature difference and 3D space difference can be established.} % Joe+Johnson
\vspace{-1mm}   
    \subsection{Local Gradient-based Heuristic Policy}
    \todo{\johnson{Our heuristic navigation method leverages the properties of our O-CNN features: (1) metric structure in feature space corresponding to real-world distances. (2) relative rotation. Given the closest place exemplar $I_i$ and current robot visual input $I_j$, initially, our policy orients the robot to the same angle as the target, using relative rotation estimate $r_{ij}$. Our policy then searches in the neighborhood area $\mathcal{N}$ by visiting the 9-by-9 grid centered at the original position. For each visited location in $\mathcal{N}$, feature distance $D_{ij}$ is computed to indicates how close the robot is to the target, forming a "potential field" where lower potential may be closer to the target. We call the aforementioned procedure the local searching process. Subsequently, the robot goes along the direction of descending gradient in the potential field by heading to the location with minimal feature distance. The policy alternates between the local searching process and heading to the location with minimal feature distance, and the robot incrementally gets closer to the target. Note that this navigation policy only roles as a naive method to prove the effectiveness of our feature extractor.}} % Johnson
\vspace{-3mm}
	\subsection{Implementation Details}

   We followed original lifted structured embedding {\cite{songCVPR16}} and used GoogleNet {\cite{googlenet}} as a feature extractor. %We use TensorFlow {\cite{tensorflow}} for training and testing our O-CNN. 
For training, the maximum training iterations were set to $20k$, the learning rate is set to $1e^{-5}$, and weights are initialized from the network pretrained on Place365 {\cite{Place365}}. The batch size is set to $32$ and margin parameter $\alpha$ is set to $1.0$. 
    Equirectangular omnidirectional images are resized to $384 \times 640$ in data pre-processing. We implement all codes with TensorFlow~\cite{tensorflow}. (Available at \url{http://aliensunmin.github.io/project/omni-cnn})
\vspace{-2mm}
\section{Data Collection}

To examine our method, we collect real world and virtual world dataset with different scenes. Details are as follows.
%Each dataset contains multiple different scenes. Details are listed below.

\begin{figure}[t]
\begin{subfigure}[virtual world framework]{
\includegraphics[width=0.5\textwidth]{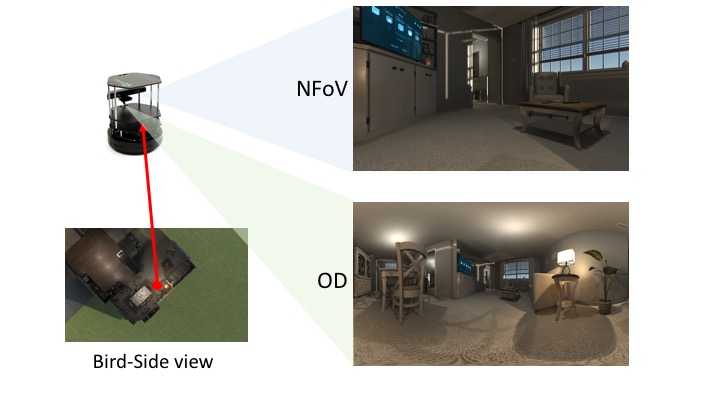}
}
\vspace{-3mm}
\label{fig:CG(a)}
\end{subfigure}

\begin{subfigure}[virtual world dataset examples]{
\includegraphics[width=0.47\textwidth]{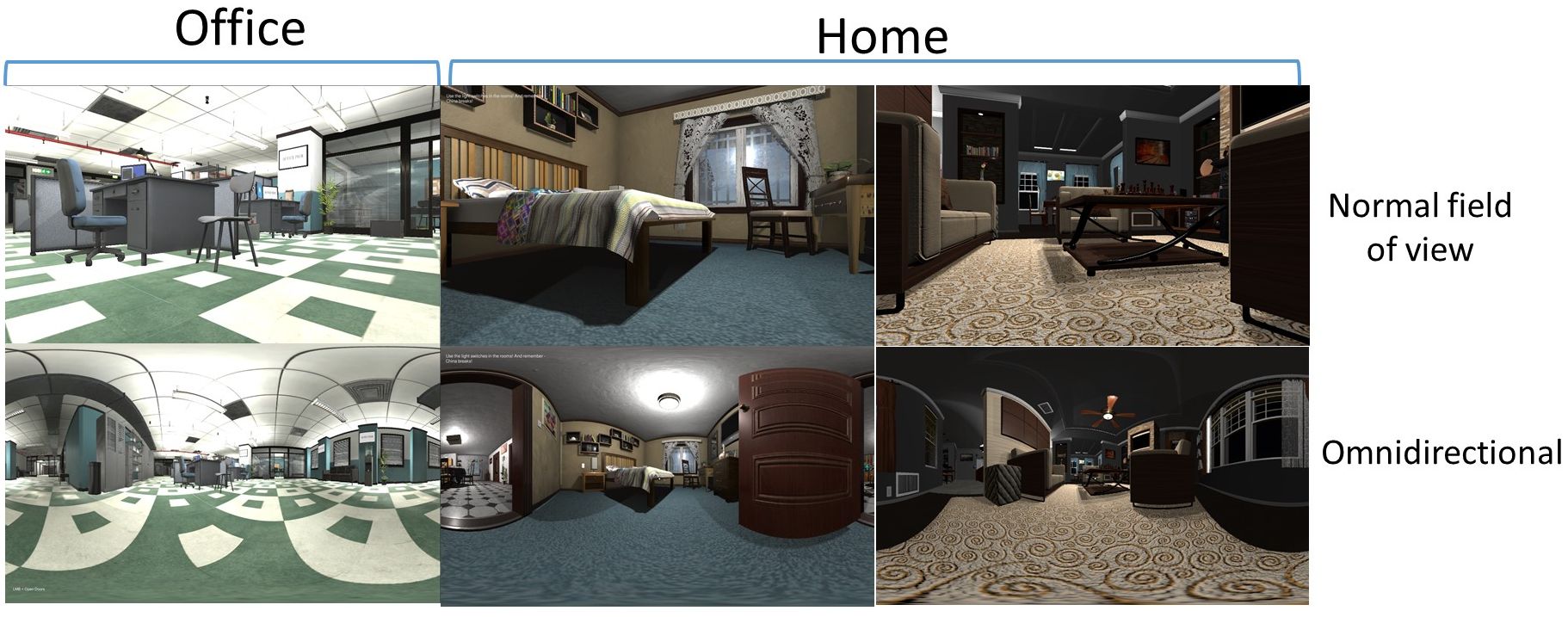}
}
\label{fig:CG(b)}
\end{subfigure}
\vspace{-3mm}
\begin{subfigure}[real world dataset examples]{
\includegraphics[width=0.47\textwidth]{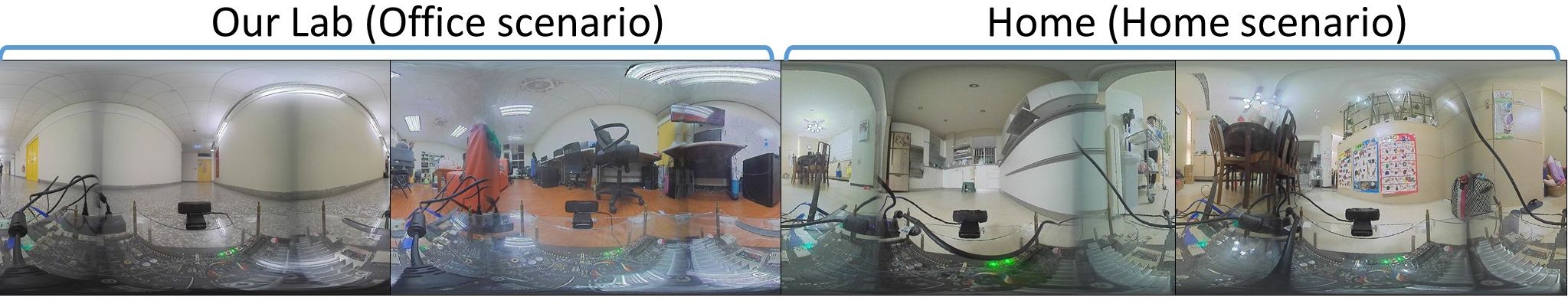}
}
\label{fig:CG(c)}
\end{subfigure}
\caption{\small \textbf{Typical examples of virtual world framework}. In (a), the top part shows the simulated normal field of view (NFoV) camera, used as training data for baselines. The bottom part shows the simulated omnidirectional (OD) camera. The bottom-left image indicates the location of our robot in the virtual environment. Both of the cameras are integrated on our robot. In (b), we show three typical examples of our virtual-world dataset. (c) Some example of real-world dataset.}
\vspace{-3mm}
\label{fig:CG_framework}
\end{figure}

\begin{table}[t!]
\centering
\resizebox{0.48\textwidth}{!}{
\begin{tabular}{|c|c|c|c|}
\hline
            & \# house scenes& \# office scenes & \# Total \\ \hline
Virtual world - training & 10    & 2                                  & 12  \\ \hline
Virtual world - evaluation & 2  & 1                              & 3  \\ \hline
Real World dataset  & 1                               & 1 & 2   \\ \hline
\end{tabular}
}
\caption{The statistics of virtual world framework and real world setting.}
%\normalsize
%\vspace{-4mm}
\label{Tab:CG_real_statistics}
\end{table}

\begin{table}[t!]
%\label{Tab:node_count}
\centering
\resizebox{0.48\textwidth}{!}{
\begin{tabular}{|c|c|c|}
\hline
   Virtual World        & Rooms per scene & Nodes per room \\ \hline
Training set/Validation set & 6/6        & 202/202  \\  \hline
Data in mapping stage/localization stage  & 6/6        &  634/830 \\ \hline
\hline
   Real World        &  Total Rooms & Total Nodes \\ \hline
   House  & 4        & 184        \\ \hline
Our Lab  & 4        & 216      \\ \hline
\end{tabular}
}
\caption{\small The statistics of collected data from CG framework and real world. \todo{\johnson{Note that the node counts are average values.}}} % Chanwei
%\normalsize
\vspace{-6mm}
\label{Tab:nodecount}
\end{table}

\subsection{Virtual World Framework}
We build our virtual environments on Unity 3D (\url{https://www.unity3d.com}), which is a game development platform with the physics engine. Popular indoor datasets for 6-DoF camera pose estimation such as 7-scene {\cite{7-scene}} were not used because of the difference of our goal and the lack of support for omnidirectional data. \todo{\johnson{Some other omnidirectional datasets, such as Eynsham dataset from FABMAP \cite{topological_cite2_fabmap}, are outdoor omnidirectional datasets, and more importantly, they don't contain robot pose labels. For these reasons, we collected our own indoor omnidirectional dataset.}} % Johnson

We collected $7$ demo scenes and some useful 3D packages from the online asset store and built another $8$ scenes based on the purchased assets. \johnson{Within the total $15$ scenes, $12$ of them are designed to emulate home scenarios, and $3$ of them are for office scenarios, shown in Table~\ref{Tab:CG_real_statistics}. For training and evaluating the proposed methods, we split the scenes into $10$ home and $2$ office scenarios as the training set and $2$ home and $1$ office scenarios as the testing set. Average room counts in a scene and average node counts in a room are shown in Table~\ref{Tab:nodecount}.} \juanting{Note that to make the virtual worlds consistent with each other, we fixed the height between the ceiling and floor to $3$ meters in each scene. Furthermore, we make sure that each room in different scenes at least has an area of $30 m^{2}$}
Fig. \ref{fig:CG_framework}(a) is an illustration of our data collection system. Fig. \ref{fig:CG_framework}(b) shows typical examples of the virtual world scenarios. 
%It is worth to note that we reshuffle the scenes in the testing set in a slightly different manner. Instead of only reshuffling small objects, we add, remove, or move larger furnitures or objects to make each scene look largely different after reshuffling. Moreover, we increase the degree of reshuffling for each time, that is the different scene having $4$ different degree of reshuffling in the testing set.

\todo{In collecting training data for place recognition, we manually picked and recorded locations from every single scene as place exemplars. For each manually-picked place exemplar, we randomly sample 9 other locations around the place exemplar and add them to the place exemplar set~\footnote{We discard the invalid images such as the images within the wall or the images basing on different altitude.}, and thus making the place exemplar set 10 times larger. \wtf{Different from the original lifted structured embedding, our continuous lifted structured embedding requires continuous labels. As a result, instead of just sampling $9$ close locations around the selected locations, we further pick $4$ additional locations which are not close to the picked locations to form place exemplars specifically for O-CNN training.}} All baseline methods (\hyperref[sec:result]{section 5}) require Normal Field-of-View (NFoV) images for training instead of our omnidirectional (referred to as OD) images. In order to provide fair comparison, for each location, we placed omnidirectional camera in the scene to take one omnidirectional image to train our model, and normal field-of-view camera in the scene to take four normal field of view images, which were taken by rotating at 0, 90, 180, and 270 degrees in order to train baseline methods with the orientation variations. \juanting{The four normal field of view images taken in each location are used to simulate the output of cubic projection from an omnidirectional image.}

\juanting{As mentioned in section~\ref{sec:method}, testing requires building a map (collecting place exemplars) and retrieving the closest exemplar. To build a map, we hired a student to control the robot to go through each room in test virtual world scene. Locations are recorded under 5Hz in mapping stage.} To collect test data for retrieving, we hired another student to randomly pick valid locations in the scene. The average number of nodes per room in the mapping stage is $634$ and the average number of nodes per room is $830$ in localization stage. Just as in the training data collection, each place exemplar recorded one omnidirectional image to test our method, and four normal field-of-view images to test baseline methods. See the typical examples of virtual world data in Fig. \ref{fig:CG_framework}(b). % Juan+Joe

\subsection{Real World Dataset}

For real-world data collection, we utilized a LUNA 360 camera to capture omnidirectional images in $2$ different indoor scenes. Those indoor scenes are one home scenario and one office scenario and each of them has multiple rooms. During the data collection, we mimic how the robot traverses each scene to take the photos. We set the size of the outputted omnidirectional image to $1280$ x $768$. We have 184 place exemplars and 266 visual inputs in House scene and 216 place exemplars and 396 visual inputs in Our Lab scene. We will make this dataset available for research purpose.
To clarify, for evaluation usage, the ground truth position of all place exemplar and visual inputs in the real-world dataset is obtained by the result after running SFM{\cite{opensfm}. The statistics of collected dataset are shown in Table~\ref{Tab:CG_real_statistics} and Table~\ref{Tab:nodecount}. See the typical examples of real-world data in Fig. \ref{fig:CG_framework}(c).
\section{Experimental Results}
\label{sec:result}

In this section, we show experiments on our virtual world dataset and our real-world dataset.
To show the effectiveness of our place recognition methods, we apply three metrics as described below.

\subsection{Metrics}\label{sec:metrics}
%The closest place exemplar finding ability and navigation ability of different methods are mostly concerned:
In our experiments, we allow each method to predict the closest place only once for each visual input. Initially, we define \textbf{error-tolerance} $e$. We consider the predicted closest place to be correct if it is within $e$ meters of the ground truth closest place. When $e$ is zero, the prediction is correct only if it is the ground truth closest place.

%\noindent \textbf{Error-tolerance.}
%We consider the predicted closest place to be correct if it is within $e$ meter of the ground truth closest place. When $e$ is zero, the prediction is correct only if it is the ground truth closest place.

\noindent \textbf{Recall-tolerance.} 
Recall is defined as the percentage of correct prediction among all visual inputs. We designed this metric to show the performance of our model at different error tolerances $e$.

\noindent \textbf{Recall-distance.} 
We fixed the error tolerance $e$ at 0.5 meters.
Then, we calculated the recall of visual inputs at a specific range of distance away from the ground truth closest place.
In particular, recall is defined as the percentage of correct predictions among visual inputs which are $m$ to $M$ meters away from the ground truth closest place.
%This metric is designed to evaluate the ability to find closest place exemplar when current visual input is in different distance from the closest exemplar. We show the graph of precision while changing the distance between current position and the closest exemplar position. The precision is defined under a fixed error tolerance 0.5 units.

\noindent \textbf{Local navigation time step.} To show the navigation ability of different methods, we report the average number of steps each method takes to navigate from a random place to its closet place.

We also conducted an ablation study to show the effectiveness of each component in our model.
%In ablation study, we adopt three version of our method:
\subsection{Ablation Study}

\noindent \textbf{O-CNN L.} Simple combination of GoogleNet {\cite{googlenet}} and \textbf{L}ifted structure embedding {\cite{songCVPR16}}.

\noindent \textbf{O-CNN LC.} We added \textbf{C}ircular padding to make the network fully utilize the information provided by the omnidirectional images.

\noindent \textbf{O-CNN LCR.} We added \textbf{R}oll branching component to make the network invariant to perspective changes introduced by the rotations in the same location.

\noindent \textbf{O-CNN CLCR.} In this version, we use the \textbf{C}ontinuous \textbf{L}ifted structure feature embedding to substitute the lifted embedding and keep the circular padding and roll branching components.

%Finally, we demonstrate the performance of our model on real robot in the real world environment.

\begin{figure*}[t!]
\centering
\includegraphics[width=0.85\textwidth]{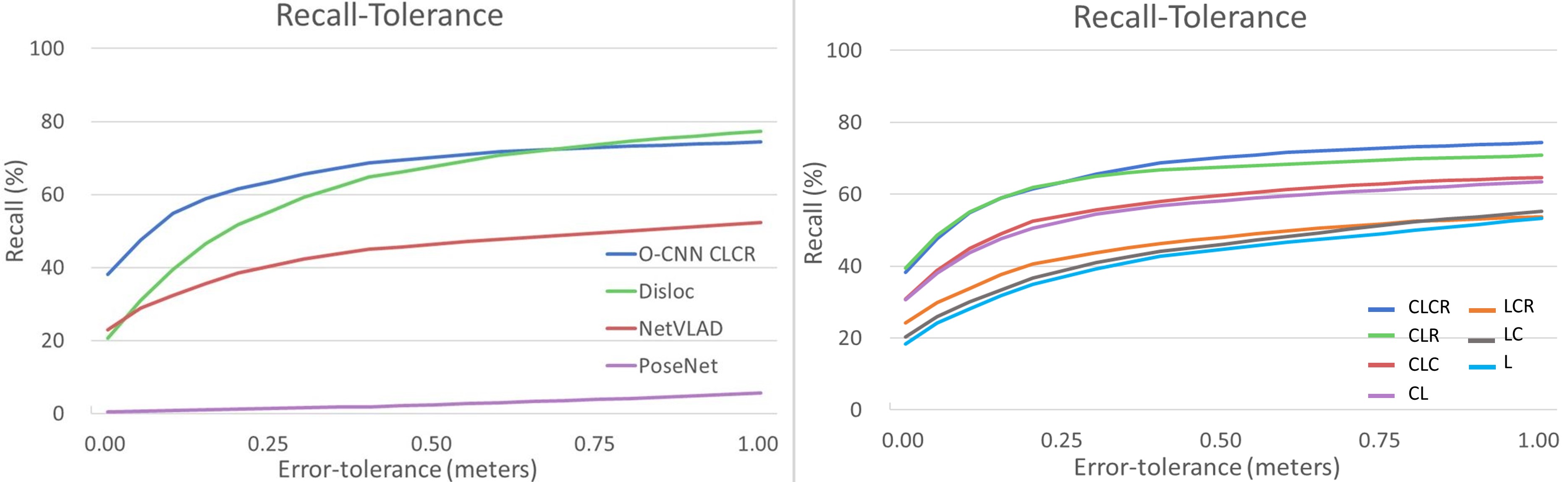}
\vspace{-2mm}
\caption{\small\textbf{The Recall-tolerance result in the virtual world.} On the left plot, we report the result compared with the baseline with the error tolerances up to 1 meter. On the right plot, we provide the result of ablation study. \todo{\johnson{Both \textbf{C}ontinuous \textbf{L}ifted loss and \textbf{R}oll branching layer bring considerable improvement in our task.}}} % Johnson
\label{fig:Recall_tolerance}
\vspace{-2mm}
\end{figure*}

\begin{figure*}[t]
\centering
\vspace{1mm}
\includegraphics[width=0.85\textwidth]{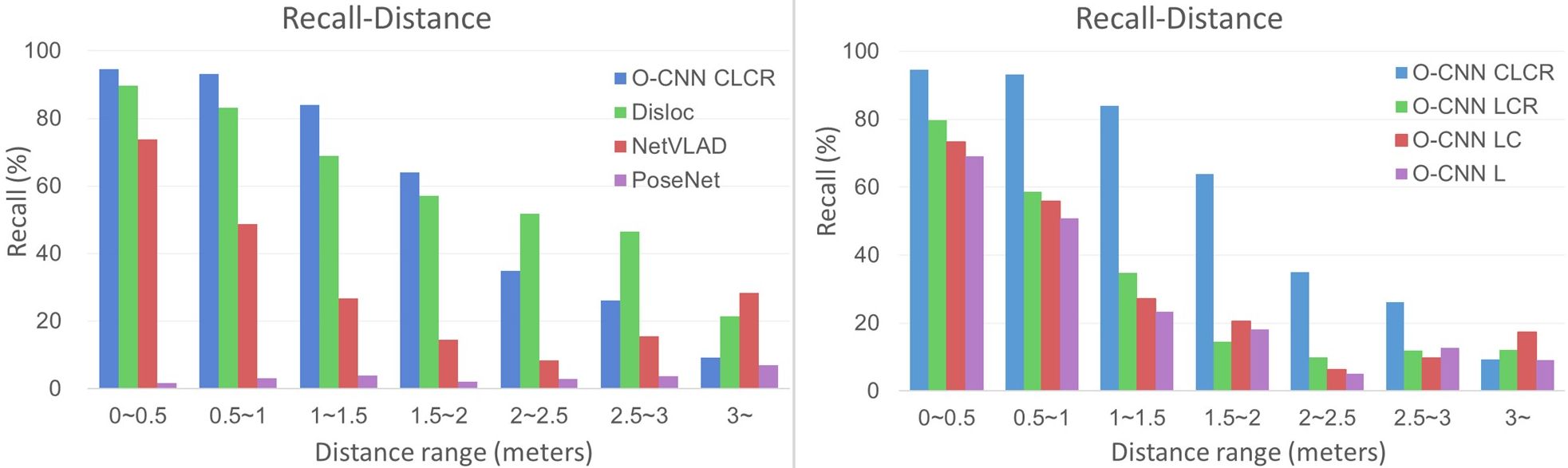}
\caption{\small \textbf{The Recall-Distance plot in the virtual world.} The left one is our method compared with baselines. The right one is the ablation study.}
\vspace{-2mm}
\label{recall_distance}
\end{figure*}

\begin{figure*}[t!]
\centering
\includegraphics[width=0.85\textwidth]{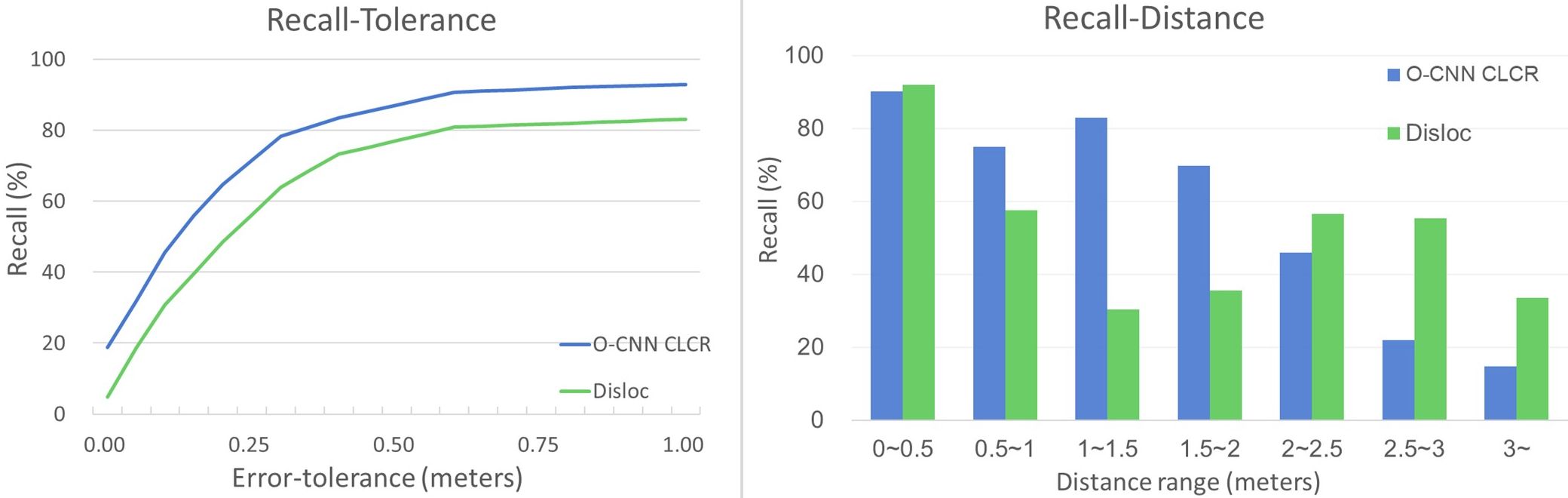}

\caption{\small \textbf{The result in real world dataset.} The left figure is the Recall-Tolerance plot. The right figure is the Recall-Distance plot.} %\todo{omit 2-3m parts?!}
\vspace{-5mm}
\label{recall_real}
\end{figure*}

We further compared our model to the following baseline models. 

\subsection{Baselines}

We compared our approach against three baselines. Image retrieval methods were chosen because they share a similar goal to ours. CNN methods are chosen because of the similarities to our learning based method. We made some minor changes to adapt baselines to our evaluation metric.

\noindent \textbf{Disloc.} Disloc \cite{disloc2014} is a state-of-the-art traditional SIFT-based method using Bag-of-Word (BoW) paradigm and Hamming embedding for image retrieval. We further improve Disloc by applying geometric burstiness proposed in {\cite{geo_bur}}, which re-rank photos by clustering them into places based on their geo-tags. To adapt Disloc to our task, we built an image vocabulary from all NFoV images. It is worth mentioning that to fairly compare the methods, four NFoV visual inputs were taken in a single location because it's common that equirectangular image to be converted to a cubic map. We abandoned the upper and lower images in the cube map and took the remaining four images as our NFoV visual inputs. Our Disloc variant then returned four target node predictions based on the four current visual inputs and applied mode estimation to merge the results. Also, Disloc is unable to be evaluated on the local navigation time step metric due to its image representation does not have a concept of distance.

\noindent \textbf{NetVLAD.} NetVLAD {\cite{NetVLAD}} is a CNN-based model. This method incorporates an encoding approach called VLAD {\cite{VLAD}}. A differentiable pooling layer which mimics the VLAD encoding is used to replace the last pooling layer of a typical CNN model. Recall that the key idea of VLAD is to match a descriptor to its closest cluster in vocabulary. For each descriptor, store the sum of the differences between the descriptors belongs to the cluster and the centroid of the cluster. To fit our problem setting, we retrain NetVLAD on our own omnidirectional images, and then directly takes omnidirectional visual input to obtain a retrieved image from mapping stage set. We compare our method with this baseline following the same evaluation metrics described in \ref{sec:metrics}.

\noindent \textbf{PoseNet.} PoseNet {\cite{posenet}} is a CNN-based 6DoF camera pose estimation method. It directly regresses camera pose given a current visual input. We end-to-end trained PoseNet on all exemplars in the map, which follows the instruction of original paper. Same as Disloc, four current visual inputs are fed into PoseNet. By assigning camera pose to the closest node, and by further applying mode estimation, we obtain the final closest exemplar prediction result.
\vspace{-1mm}
%\begin{figure}[t]
%\includegraphics[width=0.5\textwidth]{figs/Precision_error_0_1.jpg}
%\vspace{1mm}
%\includegraphics[width=0.5\textwidth]{figs/Precision_error_ablation_0_1.jpg}
%\vspace{-5mm}
%\caption{\small The Recall@1-tolerance result in the virtual world: On the upper plot, we report the result compared with the baseline with the error tolerances up to 1 meter. On the bottom plot, we provide the result of ablation study. The value of these two figures can be referred to the table 1 in the supplementary.}
%\vspace{-8mm}
%\label{fig:Precision_error}
%\end{figure}

\subsection{Main Result in Virtual World Dataset}

Here we show the experiments on the virtual world datasets described in section 4. 
\todo{\johnson{In turns of efficiency, the inference speed of one query image is $123.46$ ms, $128.21$ ms, $2439.02$ ms in O-CNN, NetVLAD, and Disloc in average.}} % Chanwei, 8.1/7.8/0.41 fps
We report the other performance measure below.
%\begin{enumerate} 

\noindent\textbf{Recall-tolerance.} According to left panel in the Fig. \ref{fig:Recall_tolerance}, we showed that our method has better performance than two deep learning based methods especially when the tolerance is small. Moreover, our method outperforms Disloc \cite{disloc2014} at error tolerance lower than 0.5 meters and takes much less inference time \todo{\johnson{(128.21 ms v.s. 2439.02 ms).}} %Chanwei, 7.8/0.41 fps
This indicates that our method is more robust and practical in application. Our method achieves best average recall rate across different error-tolerance ($69.7$\%) and the average recall of other baselines are $62.0$\%, $43.8$\%, and $2.7$\% for Disloc, NetVLAD, and PoseNet, respectively. \todo{\johnson{On the other hand, for ablation study (the right panel in Fig.~\ref{fig:Recall_tolerance}), O-CNN CLCR reaches the best performance. Additional ablation studies (CLR, CLC, CL) are provided for better illustration. As shown in the plot, the \textbf{C}ontinuous \textbf{L}ifted feature embedding (CL), \textbf{R}oll branching layer (R), and \textbf{C}ircular padding (C) bring different degree of improvements to our task.}} % Johnson
%In figure \ref{Recall-N}, we also compare the place recognition performance with baselines. Our O-CNN CLCR outperforms other methods.
%For PoseNet, it's a CNN-based architecture which takes a current visual input in the environment and directly regresses the 6-DoF camera pose. In training stage, four visual inputs converted from equiretangular image will be fed in, which the four vision inputs have different viewing-angle, this may confuse PoseNet for re-localization.

\noindent\textbf{Recall-distance.} According to the left panel in Fig. ~\ref{recall_distance}, our method outperforms other baselines when all exemplars are no larger than 2 meters away from the visual inputs. \todo{\johnson{Note that PoseNet performing extremely bad. PoseNet did conduct experiments in an indoor scene and achieved reasonable performance. However, the indoor scene is a much smaller room than scenes in our dataset.}} % Joe
Moreover, for ablation study (the right panel in Fig. ~\ref{recall_distance}), O-CNN CLCR reaches the best performance consistently. Hence, all our designed components are important. \johnson{Note that we do not show additional ablation study as those in Recall-tolerance due to space limit.}
%Furthermore, averaging across all distance range, our method's mean recall, $58.0$\%, is comparable with Disloc's, $59.8$\%. 
% In this metric, we examine how good the proposed method's ability to find closest exemplar when the current visual input is far from every exemplar in the pre-built topological metric map.
% On the contrary, the performance of NetVLAD~\cite{NetVLAD} and PoseNet~\cite{posenet} dropped when the distance is larger than 0.5m.

\noindent\textbf{Local navigation time step.} Different features are used in our heuristic policy to navigate robot. We compare O-CNN and NetVLAD {\cite{NetVLAD}}. Disloc is not compared since it is significantly slower during inference. Note that heuristic policy in NetVLAD omits information about relative rotation, which can be only derived from our method. Each episode of navigation is considered as a success if robot reaches target node under tolerance 0.3m in 100 steps. From the table \ref{Tab:local_navi_time_CG}, we can see that on average, our method outperforms NetVLAD in both success rate and speed. 

\begin{table}[t]
\centering
\small
\vspace{1.5mm}
\begin{tabular}{|c|c|c|}
\hline
           & Success rate & Avg. steps \\ \hline
Our Method & 84\%        & 22.01  \\  \hline
NetVLAD  & 68\%        &  36.14 \\ \hline
\end{tabular}
\vspace{-1mm}
\caption{\small The results for local navigation in the virtual world. The Success rate means the average successful trials and the Avg. steps means the average actions had to be taken to reach the target.}
\vspace{-7mm}
\label{Tab:local_navi_time_CG}
\normalsize
\end{table}

%\begin{figure}[t]
%\centering
%\includegraphics[width=0.5\textwidth]{figs/Distance_accuracy.jpg}
%\includegraphics[width=0.5\textwidth]{figs/Distance_accuracy_baseline.jpg}
%\caption{\small The distance-precision plot in the virtual world. The left one is our method compared with baselines. The right one is the ablation study. The value of these two figures can be referred to table 2 in the supplementary.}

%\label{distance_accuracy}
%\end{figure}

%\begin{figure}[t]
%\centering
%\includegraphics[width=0.5\textwidth]{figs/Recall-N.png}
%\caption{\small We compares place recognition performance of our method to the baselines. Note that our method outperforms the baselines with our virtual world dataset.}

%\label{Recall-N}
%\end{figure}

\subsection{Main Result in Real World Dataset}

Here we present our experimental results on real-world data. The real world experiments are only tested on O-CNN and Disloc because of their superior performance in the virtual world. Due to the limitation of collecting large-scale training data in the real world, we directly use O-CNN pre-trained on virtual world to test in the real world, and also leave local navigation time metric in the real world to future work.

%Due to the size of the room, our real-world dataset has much less testing samples than in virtual-world dataset.

\noindent\textbf{Recall-tolerance.}  According to the left panel of Fig. \ref{recall_real}, we can find that our method still outperforms the Disloc in the real world when error tolerance is low.
%and has little performance gain compared to results in virtual world.  We guess that it is because real-world images are more content-rich and the dissimilarity between images in 2 different locations may increase. Hence, our representation becomes more discriminative. 
To sum up, our method achieves better average recall across error tolerance ($77.1$\%) than Disloc ($65.4$\%).

\noindent\textbf{Recall-distance.}  
According to the right panel of Fig. \ref{recall_real}, our method is also better than Disloc in distance range no larger than 2m. Furthermore, averaging across all distance range, our method's mean recall, 57.2\%, is also larger than Disloc's, 51.5\%. 
%\todo{why Disloc getting much better in 2-3m? $\rightarrow$ there are few examples in 2-3m, do we need to omit 2-3m parts?!}

%\noindent\textbf{Local Navigation Time Step:}

\section{Conclusion}

\label{sec:conclusion}
In this paper, we have two main contributions : (1) We present a novel Omnidirectional Convolutional Neural Network (O-CNN) to improve visual place recognition and a heuristic policy to navigate a robot to the closest place on the map. (2) We propose a new virtual world framework containing lots of indoor scenes. As the experiments show, O-CNN outperforms previous learning-based methods such as NetVLAD, PoseNet, and non-learning-based method such as DisLoc. \johnson{However, there are a few limitations of the proposed method, such as the way we retrieve the closest exemplar may be computationally inefficient and how we can generalize our O-CNN features to various scenarios. We will mitigate those problems in our future works and apply our algorithm to real-world robot navigation.}

\noindent\textbf{Acknowledgement.}
We thank Institute of Information Science at the Academia Sinica, MOST-107-2634-F-007-007 for their supports. We thank Tsu-Ching Hsiao for the construction of virtual environments. Also, we thank Ching-Ju Cheng, Shun-Chang Zhong, Chung-Chih Huang, Fang-I Hsiao, Ching-Mao Chen, and Li-Yeou Wang for refining our virtual scenes.

\bibliography{reference}
\bibliographystyle{plain}
\normalsize

\end{document}